\documentclass[conference]{IEEEtran}
\IEEEoverridecommandlockouts

\usepackage{soul}
\usepackage{cite}
\usepackage{amsmath,amssymb,amsfonts}
\usepackage{algorithmic}
\usepackage{graphicx}
\usepackage{textcomp}
\usepackage{xcolor}

\makeatletter

\DeclareRobustCommand*{\RWWauthorrefmark}[1]{\raisebox{0pt}[0pt][0pt]{\textsuperscript{#1}}}%

\makeatother

\def\BibTeX{{\rm B\kern-.05em{\sc i\kern-.025em b}\kern-.08em
    T\kern-.1667em\lower.7ex\hbox{E}\kern-.125emX}}

\begin{document}

%
%
\title{TACS-RF: Tool-driven Agentic Circuit Synthesis of RF and mmWave Modules }

%
%

\author{
\IEEEauthorblockN{
Tristan Liang\RWWauthorrefmark{1},
Aravind Nagulu\RWWauthorrefmark{1}
}
\IEEEauthorblockA{
\RWWauthorrefmark{1} Department of Electrical and Computer Engineering, Northeastern University, Oakland Campus, CA 94613\\ 
\RWWauthorrefmark{1} a.nagulu@northeastern.edu
}
}

\maketitle

\begin{abstract}
We present the framework of TACS-RF --- a Tool-driven Agentic Circuit Synthesis of RF and mmWave modules. In this framework, a large-language-model (LLM) coding agent (Claude Opus 4.8) directly drives industry-standard electronic-design-automation (EDA) tools such as Cadence Virtuoso/Spectre, Calibre, Cadence EMX, and Ansys HFSS to design RF and mmWave modules. We validate the framework through the agentic design of a 240--251\,GHz sub-THz transmitter in a 90\,nm SiGe BiCMOS process. The agent semi-autonomously designed the sub-blocks including a 60\,GHz push-push Colpitts VCO with second-harmonic extraction, multistage differential cascode power drivers, a push-push frequency doubler, a differential bowtie antenna, and all the interstage matching networks. The transmitter chain with parasitic-extracted devices and EM-verified passives delivers +5.1 to +6.3\,dBm across an 11\,GHz
varactor-only tuning range, phase noise of $-$80 to $-$88\, dBc/Hz at 1\,MHz offset while consuming a DC power of 361 mW. We describe the agent-driven design flow and the resulting division of labor between human and agent, taking a step toward a ``text-to-ASIC" future for designing RF/mmWave hardware.
\end{abstract}


\section{Introduction}

 Artificial Intelligence (AI) has rapidly reshaped how humans interact with information, and is increasingly expected to reshape scientific and engineering practice. In radio-frequency (RF) and millimeter-wave/sub-terahertz integrated circuit design specifically, algorithmic and AI-assisted approaches such as deep-learning-enabled inverse design of multi-port passives and integrated circuits~\cite{karahan_natcomm_2024} and pixelated, multilayered topology-optimization frameworks for
millimeter-wave power amplifiers~\cite{chenna_jssc_2026} have demonstrated that design automation can compress schematic and layout iteration cycles from weeks to days, while achieving state-of-the-art performance without hand-crafted templates. However, these circuit-design reasoning
methods are fundamentally \emph{contextless}, each new design target requires the underlying optimization to be re-run largely from scratch, without prior topology knowledge, circuit-design reasoning or engineering heuristics from prior work.

\begin{figure}[!t]
\centering
\includegraphics[width=\linewidth]{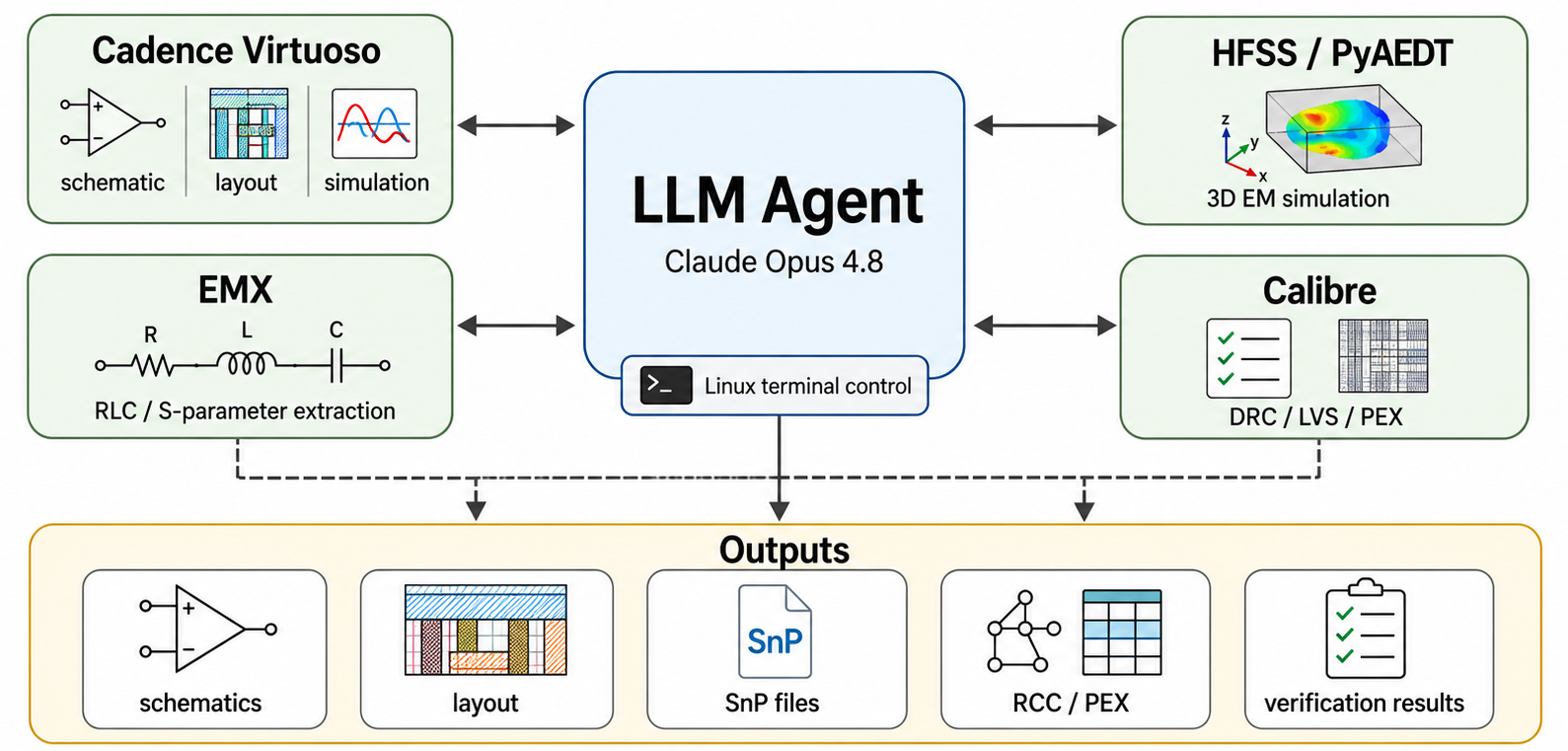}
\caption{An LLM-driven EDA workflow for RF and mmWave circuit design that integrates Cadence Virtuoso, EMX, HFSS/PyAEDT, and Calibre.}
\label{fig:AI_flow}\vspace{-20pt}
\end{figure}

\begin{figure*}[h]
\centering
\includegraphics[width=\linewidth]{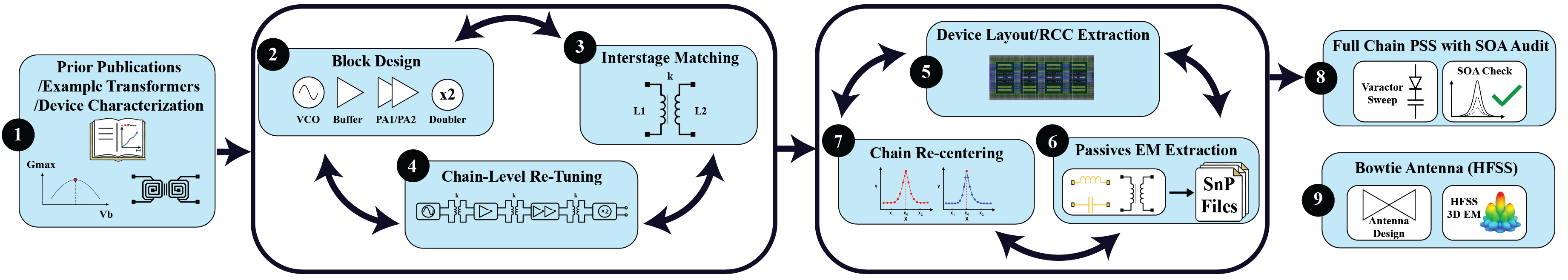}
\caption{Design flow of the sub-THz transmitter, demonstrating semi-autonomous design by the LLM agent.}\vspace{-18pt}
\label{fig:design_flow}
\end{figure*}

Large language models (LLMs), in contrast, are distinguished by their ability to reason over contextualized, cumulative knowledge such as retrieving relevant topologies, applying design equations, and adapting prior solutions to new specifications rather than re-discovering them numerically. Early studies have begun to demonstrate this potential for analog/mixed-signal design.
Ref.~\cite{circuit_agent_2025} used multiple agent frameworks to automate operational-amplifier and analog-to-digital-converter netlist generation from natural-language specifications, decomposing the design task across specialized agents for topology selection, parameter extraction, and SPICE-level verification. Separately, LLM-based sizing agents integrated in a closed loop with circuit simulators have been shown to iteratively converge the design on analog OTA and digital gates~\cite{liu_newcas_2025}. However, they do not address the EM- and parasitic-aware co-design that dominates design closure at RF, mmWave and sub-THz frequencies.

In this work, we present \textbf{TACS-RF}: \textbf{T}ool-driven \textbf{A}gentic \textbf{C}ircuit \textbf{S}ynthesis of \textbf{RF} and mmWave modules, an LLM-agent-driven framework that accelerates RF/mmWave/sub-THz module design with contextual awareness from prior publications and design iterations, while interacting with industry-standard tools (see Fig.~\ref{fig:AI_flow}). The tasks performed by the TACS framework include: (i) schematic generation and optimization, (ii) passives/antenna layout generation and EM simulations, (iii) transistor layout and parasitic extraction, and (iv) reliability and final checks. These software tools used in the workflow are directly controlled by the agent through terminal with no human-GUI interfacing. The design prototype validating the TACS framework is a sub-THz (240--251~GHz) transmitter achieving an output power of +5.1 to +6.3\,dBm and phase noise $-$80 to $-$88 dBc/Hz at 1 MHz offset across the tuning range. Through TACS-RF, design time and human labor can be reduced from months to a few days. Beyond this manuscript, we envision a ``text-to-ASIC" future where RF/mmWave circuit design, layout, and optimization tasks are reduced to prompt engineering and simulation verification.


\section{LLM--EDA Tool Integration}
\label{sec:tool_integration}

The TACS-RF framework was designed using Enterprise Claude Opus 4.8 from Anthropic as the LLM agent that directly controls the required electronic-design-automation (EDA) tools through scripts, command-line interfaces, and machine-readable simulation outputs. In this section, we describe the workflow used to connect the agent to the various software tools that constitute the design stack. The agent was provided with the design specifications, prior sub-THz circuit literature, and access to Cadence Virtuoso/Spectre, Calibre, Cadence EMX, and Ansys HFSS, and was tasked with autonomously driving each of these tools to completion, with human involvement limited to result validation, design feedback, final verification, and top-chip integration. Figure~\ref{fig:design_flow} summarizes the resulting agent-driven design flow. 

\subsection{LLM--Virtuoso Interface}
\label{subsec:virtuoso_bridge}

The agent controlled Cadence Virtuoso using \emph{virtuoso-bridge-lite}~\cite{zhang2025virtuosobridge}, an agent-oriented interface for circuit design. The bridge provides functions for generating schematics and symbols, checking connectivity, creating simulation views, launching Spectre simulations, and drawing parameterized layouts. It also returns simulator logs, operating-point data, and waveform measurements in machine-readable form, allowing the agent to evaluate each design iteration and determine the next parameter update. In our work, two complementary Spectre-based flows were used: (i) a scripted netlist flow for rapid parallel sweeps of device sizes, bias conditions, and passive values, and (ii) a Virtuoso cellview/Maestro flow for generating schematics, symbols, testbenches, and configuration views. The former supported fast iterative optimization, whereas the latter was used for human verification.

\begin{figure}[!t]
    \centering
    \includegraphics[width=\linewidth]{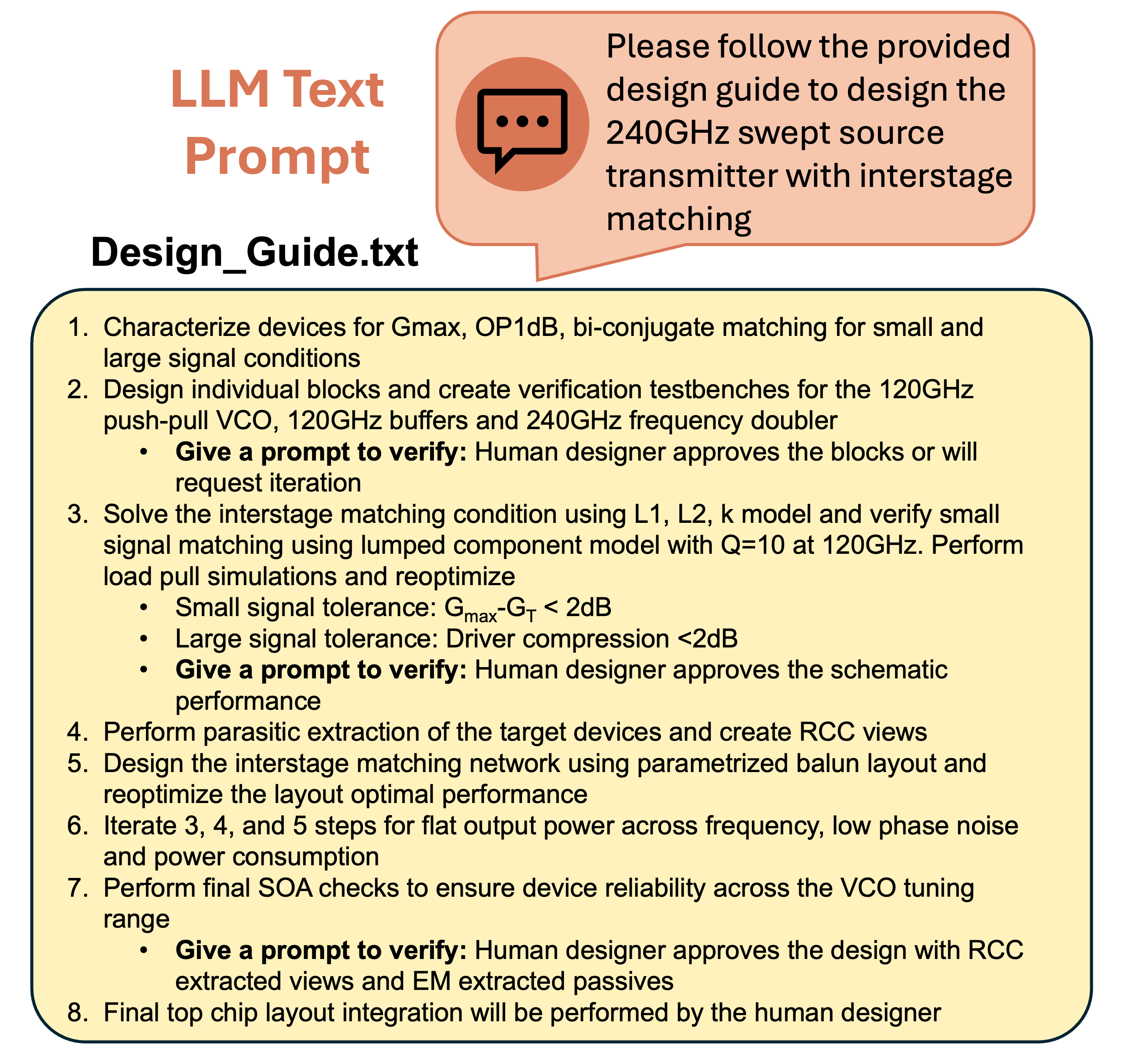}\vspace{-10pt}
    \caption{A representative design guideline provided to the LLM-agent.}\vspace{-10pt}
    \label{fig:Design_prompt}
\end{figure}

\subsection{LLM Control of Electromagnetic Simulators}
\label{subsec:em_control}

The agent used Cadence EMX to design the on-chip transformers and inductors as parameterized single-loop octagonal structures in the top two metal layers. The multiport S-parameter models of the EM simulations were passed back to Cadence Virtuoso for circuit-level evaluation. The geometry was iteratively updated based on the transmitter performance from Virtuoso simulation.

Additionally, an on-chip bowtie antenna was designed by the agent in Ansys HFSS through agent-generated PyAEDT scripts using the technology stackup and material properties. The agent generated the antenna layout, defined the ports, radiation boundaries, meshing, and frequency sweep, and exported the resulting matching, gain, and directivity data.

\subsection{Calibre-Based Physical Verification and Extraction}
\label{subsec:calibre_automation}

The agent also automated the Calibre design-rule checking (DRC), layout-versus-schematic (LVS), and parasitic-extraction flows through command-line execution and scripted runset configuration. Parameterized transistor layouts were provided by the human designer. The agent iteratively verified DRC, LVS, and extracted the resistance, capacitance and coupling-capacitance (RCC) parasitics of the transistor layout and imported the RCC-view for performance evaluation. 

\section{LLM-Designed Sub-THz Transmitter}
\label{sec:design}

\begin{figure*}[!t]
    \centering
    \includegraphics[width=\textwidth]{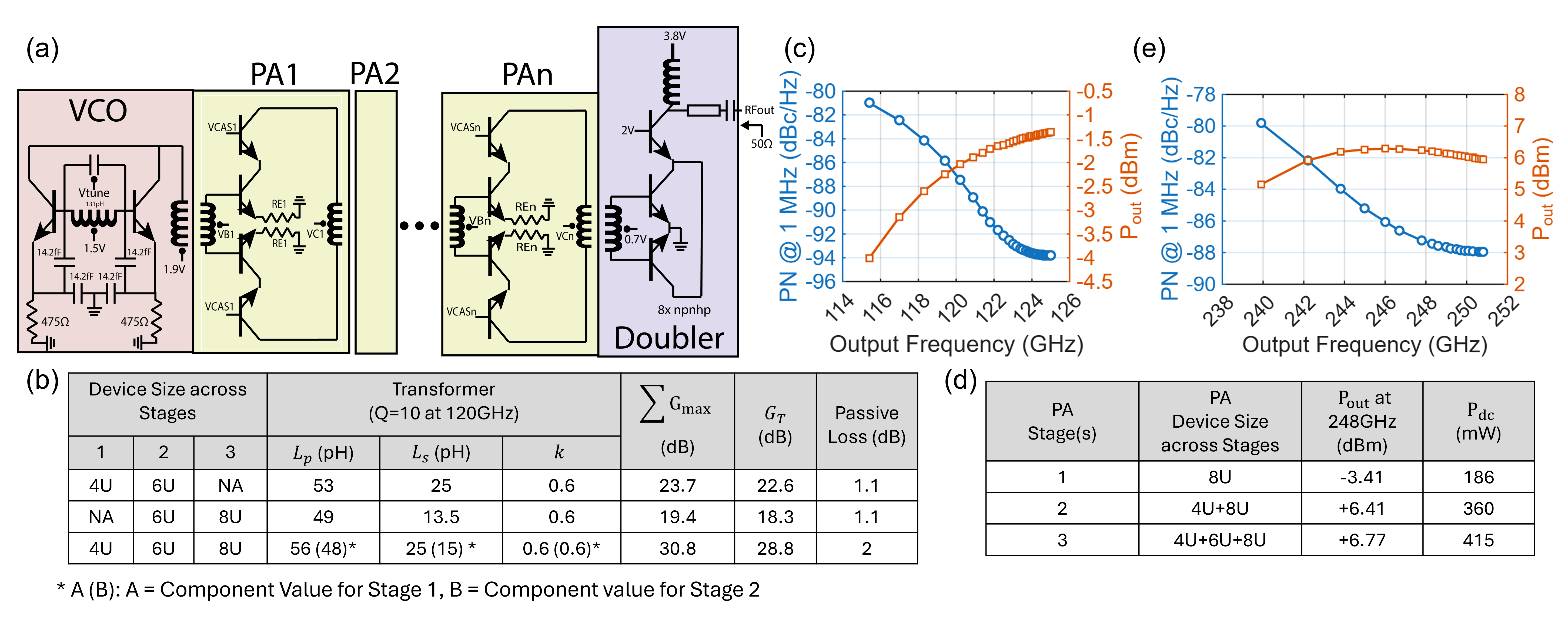}\vspace{-10pt}
    \caption{Agent-designed 240--251~GHz transmitter. (a)~Complete transmitter schematic in a 90\,nm SiGe BiCMOS process. (b)~Automated design results for the small signal inter-stage matching and achieved performance benchmarks. (c) Simulated performance of the 60\,GHz VCO with second harmonic extraction. (d)~Schematic and lumped model based performance optimization of the sub-THz transmitter across varying number of driver stages. (e)~Simulations of the transmitter EM extracted passives and RCC-extracted devices: output power ($P_{out}$) and phase noise (PN) at a 1~MHz offset across the varactor $V_{tune}$ sweep.}\vspace{-15pt}
    \label{fig:Schematic}
\end{figure*}

The target design is a swept-frequency sub-THz transmitter for applications including high-precision radar and gas spectroscopy. Fig.~\ref{fig:design_flow} depicts the adopted design flow emulating a typical human-design flow with two iterative optimization loops for schematic-only and layout-extracted parasitics. Fig.~\ref{fig:Design_prompt} shows a representative design guide provided to the agent. Fig.~\ref{fig:Schematic}(a) depicts the system architecture, the signal chain consists of a 60\,GHz push-push Colpitts VCO with second-harmonic extraction, multiple stages of differential 120\,GHz gain, a push-push frequency doubler, and an on-chip bowtie antenna operating near 240\,GHz. 

\subsection{Technology and Reference-Design Setup (Step 1):}
The agent was provided with selected prior sub-THz transmitter publications~\cite{Schmalz_RFIC2012,Shopov_JSSC2016}, target specifications, and a set of example transformer layouts. Using virtuoso-bridge-lite, the agent characterized the SiGe NPN BJTs with different multiplicities of the unit transistor (U: L=2\,$\mu$m and cbebc configuration) and recorded $G_{\max}$, $f_{\mathrm{T}}/f_{\max}$, current consumption across bias conditions. From the example transformer layouts, the agent analyzed the layout strategy and also EM simulated the passives to extract the quality factor ($\approx$10 at 120\,GHz) and coupling coefficient ($<$0.7) of the interstage matching networks. Combining the reference topologies, device characterization, and passive-loss estimates, the agent selected a candidate system topology (see Fig.~\ref{fig:Schematic}(a)) and performed a system-level feasibility analysis.

\begin{figure}[!t]
    \centering
    \includegraphics[width=\linewidth]{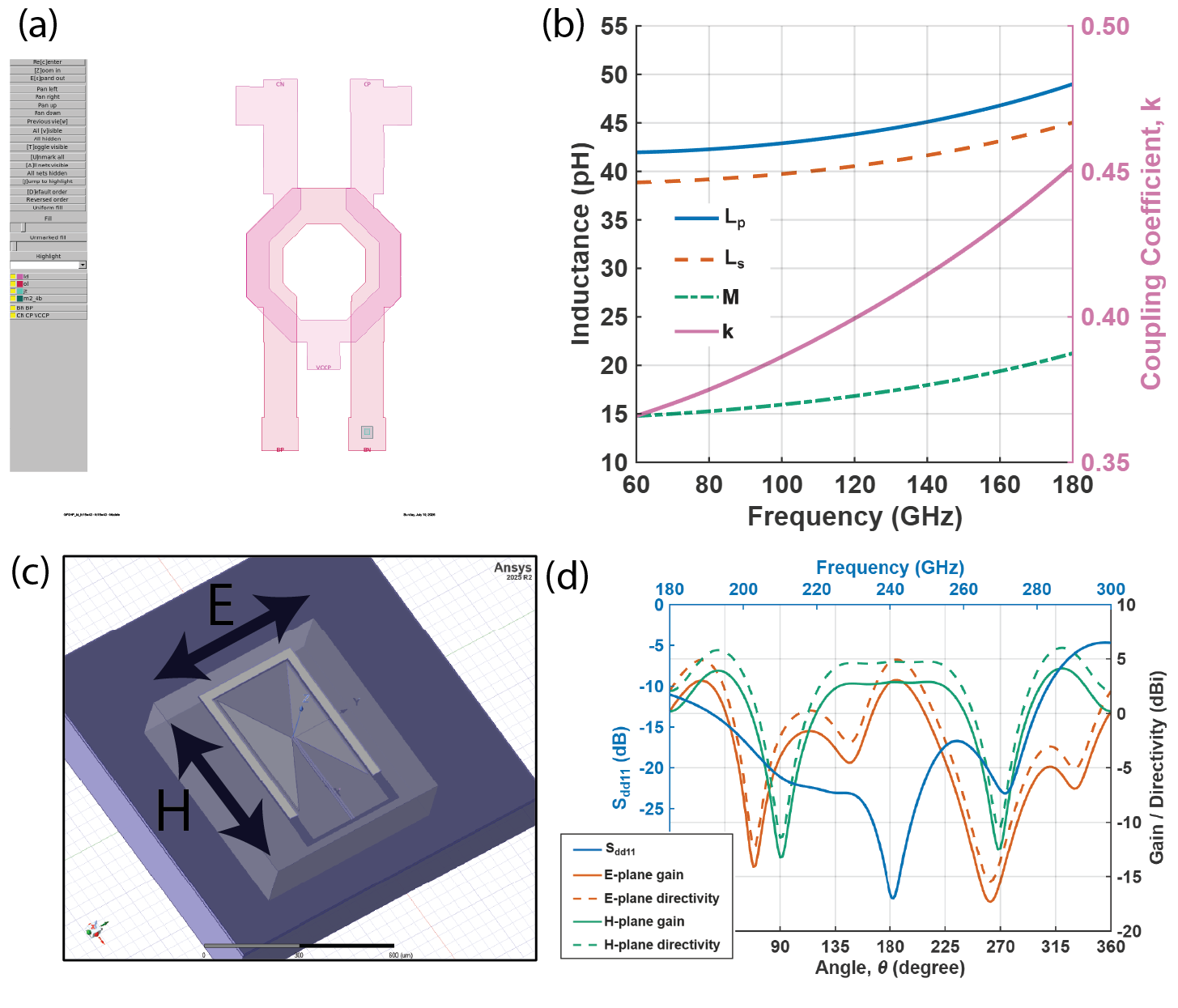}
    \caption{TACS-RF framework performing EM simulations using Ansys HFSS and Cadence EMX. A representative transformer balun design in EMX: (a) layout and (b) extracted parameters. A 240\,GHz antenna design in HFSS: (c) antenna layout, and (d) simulated input matching, gain, and directivity. }
    \label{fig:tx_antenna}
\end{figure}

\subsection{Schematic-level Circuit and Matching Design (Steps 2--4):}
Using these results, the agent optimized the device sizes and bias points of the VCO, buffer, PA stages, and frequency doubler in independent testbenches (see Fig.~\ref{fig:Schematic}(b)) and benchmarked the achieved performance (see Fig.~\ref{fig:Schematic}(c)). Finally, the number of driver stages, device dimensions, bias conditions, and matching parameters were iteratively co-optimized to establish the best achievable schematic-level output power, tuning range, and phase-noise performance. The matching networks during this design phase are represented as parametrized transformers with achievable quality factor and coupling coefficient. It can be noticed from Fig.~\ref{fig:Schematic}(d) that two driver stages are sufficient for this design, while a third driver stage can be adopted to provide additional power-gain margin, offering more flexibility to absorb post-layout losses due to inter-stage detuning.

\begin{table*}[!t]
\centering
\caption{Comparison With Prior Sub-THz SiGe Sources}
\label{tab:comparison}
\resizebox{\textwidth}{!}{%
\begin{tabular}{l c c c c c c}
\hline
\textbf{Reference} & \textbf{Process} & \textbf{$f_T/f_{max}$} & \textbf{Frequency} & \textbf{$P_{out}$} & \textbf{$P_{dc}$} & \textbf{Phase Noise} \\
 & & \textbf{(GHz)} & & & & \\
\hline
\cite{Schmalz_RFIC2012}$^{\mathrm{a}}$ & 0.13~$\mu$m SiGe & 300/500 & 229--251~GHz & 1--2~dBm & 0.29~W & $-$85~dBc/Hz @ 1~MHz \\
\cite{Shopov_JSSC2016}$^{\mathrm{b}}$ & 55~nm SiGe & 330/350 & 234--261~GHz & 5.4--7.2~dBm & 0.38~W & $-$105~dBc/Hz @ 10~MHz \\
\textbf{This work}$^{\mathrm{c}}$ & 90~nm SiGe & 300/350 & 240--251~GHz& 5.1--6.3~dBm & 0.36~W & $-$88$^\mathrm{d}$ @ 1~MHz; $-$108$^\mathrm{d}$ @ 10~MHz \\
\hline
\multicolumn{7}{l}{\footnotesize $^{\mathrm{a}}$ Push-push VCO + PA + doubler. $^{\mathrm{b}}$ 120~GHz VCO + LO tree + doubler. $^{\mathrm{c}}$ Simulated -- Push-push VCO + 2 PAs + doubler. $^\mathrm{d}$ operating at 248\,GHz.} \vspace{-10pt}
\end{tabular}%
}
\end{table*}

\subsection{Layout- and EM-aware Optimization (Steps 5--7):}
A library of transistor layouts covering the required device sizes was prepared by the human designer. The agent selected the corresponding layouts, executed Calibre LVS and RCC extraction, and replaced the schematic transistors with their extracted models. In parallel, the interstage transformers were parameterized using the top two metal layers and simulated in EMX (see Fig.~\ref{fig:tx_antenna}(a) and (b)). Their multiport S-parameter files were imported into Virtuoso to replace the ideal matching networks. The agent iteratively adjusted the circuit and passive parameters until the complete chain was re-centered over the target 240\,GHz band and optimized for output power.

The overall layout and electromagnetic simulation optimization loop was $\text{RCC extraction}\rightarrow \text{passive geometry generation}\rightarrow\text{EM simulation}\rightarrow\text{circuit evaluation}\rightarrow\text{circuit refinement}.$ This closed-loop procedure allowed the passive structures to be optimized according to their impact on the complete transmitter rather than solely according to isolated passive metrics. Fig.~\ref{fig:Schematic}(e) summarizes the achieved performance using RCC devices and EM simulated passives. 

\subsection{Top-Level Integration and Verification (Steps 8--9):}
In Step~8, the complete extracted transmitter was simulated across the VCO tuning voltage, resulting in an operating frequency of 240--251\,GHz. The matching and device parameters were optimized for large signal matching and power delivery. Additionally, the instantaneous terminal voltages of every transistor were checked against the PDK safe-operating-area (SOA) limits across the full tuning range. In Step~9, the agent independently designed an on-chip bowtie antenna in HFSS, and extracted its input matching, gain, directivity, and radiation patterns (see Fig.~\ref{fig:tx_antenna}(c) and (d)). Finally, the human designer would assemble the RCC-extracted active blocks, EM-extracted matching structures and antenna into the complete top-level layout for final verification.

\section{Post-Extraction Simulation Results}

\label{sec:results}


\subsection{Per-Block Performance}
The standalone VCO delivers $-$1.5~dBm at 123.7~GHz into its load-pull optimum of $25 + \mathrm{j}30~\Omega$. 
The biasing of the two-stage PA can be tuned for a power gain of 14.1\,dB--16.6\,dB, with an $\mathrm{OP}_{1\mathrm{dB}}$ of +2.2~dBm, and a $P_{\mathrm{sat}}$ of +8.5~dBm into the power-optimum load. Under optimal operation, the doubler requires an input power of +2.9~dBm at 123\,GHz and delivers +6.3~dBm at 246~GHz, achieving a conversion gain of +3.4~dB.

\subsection{Final Transmitter Chain Performance}

The complete chain delivers +6.3~dBm at 246~GHz while consuming 361\,mW. The phase noise (PN) and output power ($P_{out}$) of the transmitter across the VCO tuning voltage ($V_{tune}$) sweep are plotted in Fig.~\ref{fig:Schematic}(e), resulting in a tuning range of 11\,GHz.

Finally, Table~\ref{tab:comparison} compares our post-extraction results with the prior state-of-the-art sub-THz sources that were provided to the Claude agent as reference designs. The TACS-RF based design achieves highly competitive performance against these established benchmarks.

\section{Conclusion}
\label{sec:conclusion}

Using the TACS-RF framework, a sub-THz transmitter chain with parasitic-extracted devices and EM-verified passives was designed in under a week, compared with 2–3 months for a comparable manual design flow. The primary bottleneck was EM simulation runtime; a library of pre-characterized passives would reduce design time further. With established design guidelines, prior schematics, and reference layouts as the agent knowledge base, we expect the workflow to extend to a variety of RF/mmWave modules. Because the agent iterates semi-autonomously across schematic, EM, and extraction tools, the human role shifts from execution to system definition, ideation, and verification. This can lower the barrier to entry for RF/mmWave IC design, shorten design cycles, and reduce engineering effort. As LLMs continue to improve in reasoning and coding proficiency, we expect fully autonomous ``text-to-ASIC" design to become achievable.

\section*{Acknowledgment}
This work was supported in part by NSF Award \#2529237, in part by NSF Award \#2513548, and in part by NSF Award \#2519886.

\bibliographystyle{IEEEtran}
\bibliography{references}

\vspace{12pt}

\end{document}